\documentclass[letterpaper]{article} 
\usepackage{aaai23}  
\usepackage{times}  
\usepackage{helvet}  
\usepackage{courier}  
\usepackage[hyphens]{url}  
\usepackage{graphicx} 
\usepackage{natbib}  
\usepackage{caption} 
\usepackage{algorithm}
\usepackage{algorithmic}
\usepackage{newfloat}
\usepackage{listings}
\DeclareCaptionStyle{ruled}{labelfont=normalfont,labelsep=colon,strut=off} 
\floatstyle{ruled}
\newfloat{listing}{tb}{lst}{}
\floatname{listing}{Listing}
\nocopyright 

\title{Towards Accurate, Energy-Efficient, \& Low-Latency Spiking LSTMs}
\author{
    Gourav Datta\textsuperscript{\rm 1},
    Haoqin Deng\textsuperscript{\rm 2}\thanks{Work done at University of Southern California},
    Robert Aviles\textsuperscript{\rm 1},
    Peter A. Beerel\textsuperscript{\rm 1}
}
\affiliations{
    \textsuperscript{\rm 1}University of Southern California, USA \ \ \textsuperscript{\rm 2} University of Washington, Seattle, USA \\

}

\usepackage{bibentry}

\usepackage[utf8]{inputenc} 
\usepackage[T1]{fontenc}    
\PassOptionsToPackage{hyphens}{url}\usepackage{hyperref}
\usepackage{booktabs}  
\usepackage{subcaption}
\usepackage{nicefrac}       
\usepackage{microtype}      
\usepackage{comment}
\newcommand{\centered}[1]{\begin{tabular}{l} #1 \end{tabular}}
\usepackage{xcolor} 
\usepackage{makecell}
\usepackage{amsmath,amssymb,amsfonts}
\usepackage{mathtools} 
\usepackage{amssymb}
\usepackage{pifont}

\newcommand{\floor}[1]{\left\lfloor #1 \right\rfloor}
\newcommand{\ceil}[1]{\left\lceil #1 \right\rceil}

\newcommand{\cmark}{$\checkmark$}
\newcommand{\xmark}{$\times$}
\newcommand{\rev}[1]{{\color{black} #1}}

\begin{document}

\maketitle

\begin{abstract}

Spiking Neural Networks (SNNs) have emerged as an attractive spatio-temporal computing paradigm for complex vision tasks. However, most existing works yield models that require many time steps and do not leverage the inherent temporal dynamics of spiking neural networks, even for sequential tasks. Motivated by this observation, we propose an \rev{optimized spiking long short-term memory networks (LSTM) training framework that involves a novel ANN-to-SNN conversion framework, followed by SNN training}. In particular, we propose novel activation functions in the source LSTM architecture and judiciously select a subset of them for conversion to \rev{integrate-and-fire} (IF) activations with optimal bias shifts. Additionally, we derive the \rev{leaky-integrate-and-fire} (LIF) activation functions converted from their non-spiking LSTM counterparts which justifies the need to jointly optimize the weights, threshold, and leak parameter. We also propose a pipelined parallel processing scheme which hides the SNN time steps, significantly improving system latency, especially for long sequences. The resulting SNNs have high activation sparsity and require only accumulate operations (AC), in contrast to expensive multiply-and-accumulates (MAC) needed for ANNs, except for the input layer when using direct encoding, yielding significant improvements in energy efficiency. We evaluate our framework on sequential learning tasks including temporal MNIST, Google Speech Commands (GSC), and UCI Smartphone datasets on different LSTM architectures. We obtain test accuracy of $94.75$\% with only $2$ time steps with direct encoding on the GSC dataset with ${\sim}4.1\times$ lower energy than an iso-architecture standard LSTM. 

\end{abstract}

\section{Introduction \& Related Work}\label{sec:intro}

In contrast to the neurons in ANNs, the neurons in Spiking Neural Networks (SNNs) are biologically inspired, receiving and transmitting information via spikes. SNNs promise higher energy-efficiency than ANNs due to their high activation sparsity and event-driven spike-based computation \cite{diehl2016conversion} which helps avoid the costly multiplication operations that dominate ANNs.
To handle multi-bit inputs, such as typical in traditional datasets and real-life sensor-based applications, however, the inputs are often spike encoded in the temporal domain using rate coding \cite{diehl2016conversion}, temporal coding \cite{comsa_2020}, or rank-order coding \cite{Kheradpisheh_2020}.
Alternatively, instead of spike encoding the inputs, some researchers explored directly feeding the analog pixel values in the first convolutional layer, and thereby, emitting spikes 
only in the subsequent layers \cite{rathi2020dietsnn}. This can dramatically reduce the number of time steps needed to achieve the state-of-the-art accuracy, but comes at the cost that the first layer now requires MACs \cite{rathi2020dietsnn, datta2022fin, kundu2021lowlatency}. 
However, all these encoding techniques increase the end-to-end latency (proportional to the number of time steps) compared to their non-spiking counterparts.

In addition to accommodating various forms of spike encoding, supervised learning algorithms for SNNs, such as surrogate gradient learning (SGL) have overcome various roadblocks associated with the discontinuous derivative of the spike activation function \cite{lee_dsnn, kim_2020, neftci_surg, panda_res}. It is also commonly agreed that SNNs following the integrate-and-fire (IF) compute model can be converted from ANNs with low error by approximating the activation value of ReLU neurons with the firing rate of spiking neurons \cite{dsnn_conversion_abhronilfin, rathi2020iclr, diehl2016conversion}. SNNs trained using ANN-to-SNN conversion, coupled with SGL, have been able to perform similar to SOTA CNNs in terms of test accuracy in traditional image recognition tasks \cite{rathi2020dietsnn, rathi2020iclr} with significant advantages in compute efficiency. Previous works \cite{rathi2020dietsnn, datta2021training, kundu_2021} have adopted SGL to jointly train the threshold and leak values to improve the accuracy-latency 
tradeoff but without any analytical justification.

Inspite of numerous innovations in SNN training algorithms for static \cite{panda2016_sup,panda_res,rathi2020dietsnn,rathi2020iclr,kim_2020} and dynamic vision tasks \cite{dynvis_1,dynvis_2}, there has been relatively fewer research that target SNNs for sequence learning tasks.
Among the existing works, some are limited to the use of spiking inputs \cite{rezaabad2020lstmspike, ponghiran2021seqsnn}
which might not represent several real-world use cases. Furthermore, some \cite{deng2021optimal,Moritz2019UnidirectionalNN, diehl2016rnn_snn} propose to yield SNNs from vanilla RNNs which has been shown to yield a large accuracy drop for large-scale sequence learning tasks, as they are unable to model temporal dependencies for long sequences. Others \cite{ponghiran2021hyrbidsnn} use the same input expansion approach for spike encoding and yield SNNs which requires serial processing for each input in the sequence, severely increasing total latency. 
A more recent work \cite{ponghiran2021seqsnn} proposed a more complex neuron model compared to the popular IF or leaky-integrate-and-fire (LIF) model, to improve the recurrence dynamics for sequential learning. 
Additionally, it lets the hidden activation maps be multi-bit (as opposed to binary spikes) which improves training, but requires multiplications that reduces energy efficiency compared to the multiplier-less SNN models we develop.
In particular, our work leverages both the temporal and sparse dynamics of SNNs to reduce the inference latency and energy consumption of large-scale streaming ML workloads while achieving close to SOTA accuracy.

The key contributions of our work are summarized below.

\begin{itemize}
    \item We propose a training framework \rev{that involves the conversion from a pre-trained non-spiking LSTM} to a spiking LSTM model that minimizes conversion error. Our framework involves three novel techniques. i) Converting the traditional sigmoid and tanh activation functions in the source LSTM to clipped versions, ii) judiciously selecting a subset of these functions for conversion to IF activation functions such that the SNN does not require the expensive MAC operations, and iii) finding the optimal shifts of the IF activation functions.
    \item To the best of our knowledge, we are the first to obtain a closed form expression of the LIF activation function which, in particular, captures the impact of the leak term. This function helps us analyze the post-conversion error between the non-spiking LSTM and LIF activation outputs under non-uniform and non-identical input distributions and motivates its reduction by jointly training the threshold and leak term.
    
    \item We propose a high-level parallel and pipelined implementation of the resulting SNN-based computations, 
    which coupled with our training algorithm, results in negligible latency overheads compared to the baseline LSTM and improves the hardware utilization.
    \item We demonstrate the energy-latency-accuracy trade-off benefits of our proposed framework through FPGA synthesis and place-and-route, extensive ML experiments with different LSTM architectures on sequential tasks from computer vision (temporal MNIST), \rev{spoken term classification} (Google Speech Commands) and human activity recogniton (UCI Smartphone) applications, and comparisons with existing spiking and non-spiking LSTMs.       
\end{itemize}

\section{Preliminaries}

\subsection{SNN IF/LIF Models}

In this work, we adopt the popular IF and LIF models \cite{leefin2020} to capture the computation dynamics of an SNN. In both these models, a neuron transmits binary spike trains (except the input layer for direct encoding) over a total number of pre-defined time steps. To incorporate the temporal input dimension, each neuron has an internal state called its membrane potential $U_i(t)$ which captures the integration of the weight (denoted as $W_{ij}$) modulated incoming spikes (denoted as $S_j(t)$). In the LIF model, $U_i(t)$ leaks with a fixed time constant, denoted as $\lambda$ ($\lambda=1$ for IF model). 
With the spiking threshold represented as $V^{th}$, the LIF neuron dynamics are expressed as
\begin{align}
U_i^{temp}(t)&=\lambda U_i(t-1)+\sum_j W_{ij}{S_j(t)} \\ 
S_i(t)&=
\begin{cases}
    V^{th}, & \text{if } U_i^{temp}(t)>V^{th}\label{eq:lif_output}\\
    0, & \text{otherwise} \\
\end{cases} \\
U_i(t) &= U_i^{temp}(t)-S_i(t)
\label{eq:IF_out_spike}
\end{align}

\subsection{Surrogate Gradient Learning}

Since the spiking neuron functionality is discontinuous and non-differentiable, it is difficult to implement gradient descent based backpropagation in SNNs. Hence, previous works \cite{lee_2020, neftci_surg} approximate the spiking function with a continuous differentiable function, which helps back-propagate non-zero gradients known as surrogate gradients. The resulting weight update in the $l^{th}$ hidden layer in the SNN is calculated as  
\begin{align}
    \Delta{W_l}{=}\sum_{t}\frac{\partial\mathcal{L}}{\partial W_l}{=}\sum_{t}\frac{\partial\mathcal{L}}{\partial\mbox{$O$}_l^t}\frac{\partial\mbox{$O$}_l^t}{\partial\mbox{$U$}_l^t}\frac{\partial\mbox{$U$}_l^t}{\partial W_l}{=}\sum_{t}\frac{\partial\mathcal{L}}{\partial\mbox{$O$}_l^t}\frac{\partial\mbox{$O$}_l^t}{\partial\mbox{$U$}_l^t}\mbox{$O$}_{l-1}^t \notag
\end{align}
\noexpand
where $\mbox{O}_l^t$ and $\mbox{U}_l^t$ denote the spike output and membrane potential tensor of the $l^{th}$ layer respectively at time step $t$. $\frac{\partial{O}_l^t}{\partial{U}_l^t}$ is the non-differentiable gradient which can be approximated with the surrogate gradient,  $\frac{\partial{O}_l^t}{\partial{U}_l^t}=\frac{\gamma}{V_l^{th}}\cdot{max(0,1-\text{abs}(\mbox{$\frac{U_l^t}{V_l^{th}}-1$))}}$, where $V_l^{th}$ is the $l^{th}$ layer threshold and $\gamma$ is a hyperparameter denoting the maximum gradient value \cite{bellec_2018long}.

\begin{figure*}
\centering
\begin{subfigure}{0.49\textwidth}
    \includegraphics[width = \textwidth]{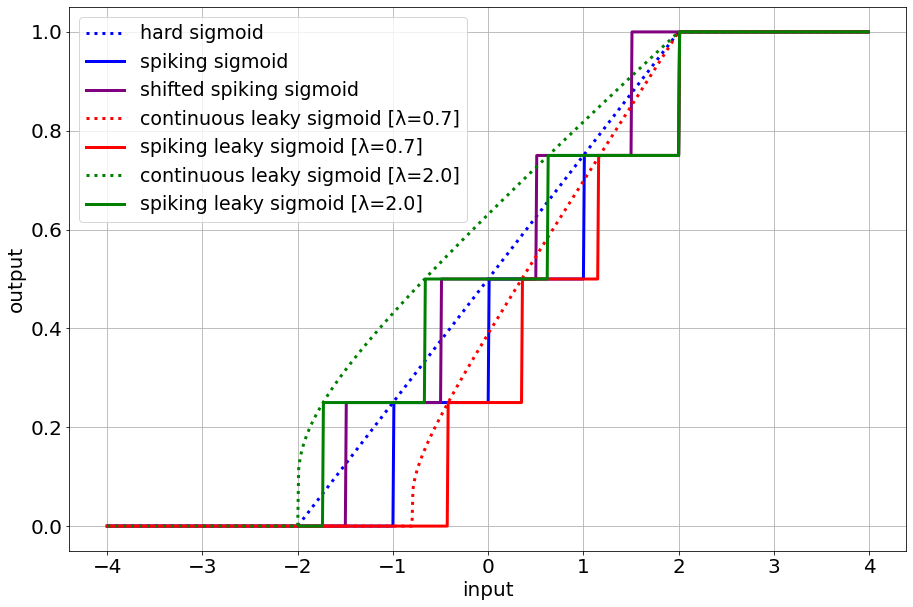}
    \caption{ }
    \label{fig:sigmoid}
    
\end{subfigure}
\begin{subfigure}{0.49\textwidth}
    \includegraphics[width = \textwidth]{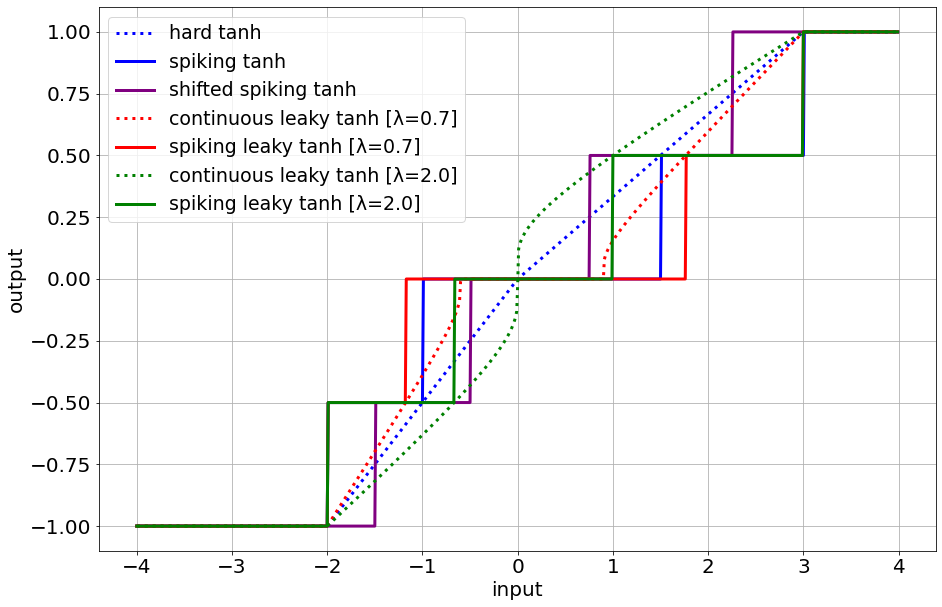}
    \caption{ }
    \label{fig:tanh}
\end{subfigure}
\hfill
\caption{(a) Hard Sigmoid and (b) hard tanh, along with their proposed bias shifts, converted IF and LIF activation functions for $4$ time steps, $V^{th}_{sig}=4$, $V^{th}_{tanh+}=3$, $V^{th}_{tanh-}=-2$. }
\label{fig:act_func}
\vspace{-2mm}
\end{figure*}

\section{Proposed Training Framework}

\subsection{Non-spiking LSTM}

In order to yield accurate LSTM-based SNN models, we first replace the traditional tanh and sigmoid activation functions in the baseline LSTM model with their \textit{hard} (clipped) versions, as illustrated in Fig. \ref{fig:act_func}(a-b). Unlike previous works \cite{ponghiran2021hyrbidsnn}, we decouple the hard tanh function into two hard sigmoid functions. Hence, we have a single threshold value, denoted as $V_{sig}^{th}$ for the hard sigmoid function whose outputs are always positive, but two threshold values (one positive, denoted as $V_{tanh+}^{th}$ for output values ranging from 0 to +1 and one negative denoted as $V_{tanh-}^{th}$ for output values ranging from 0 to -1) for the hard tanh function. This approach enables both the hard sigmoid and tanh functions to be implemented with threshold ReLU functions which have been shown to improve the accuracy of ANN-to-SNN conversion \cite{dsnn_conversion_abhronilfin, deng2021optimal}.

\subsection{Conversion to SNN}

\begin{figure*}[t!]
\begin{center}
\includegraphics[width=\textwidth]{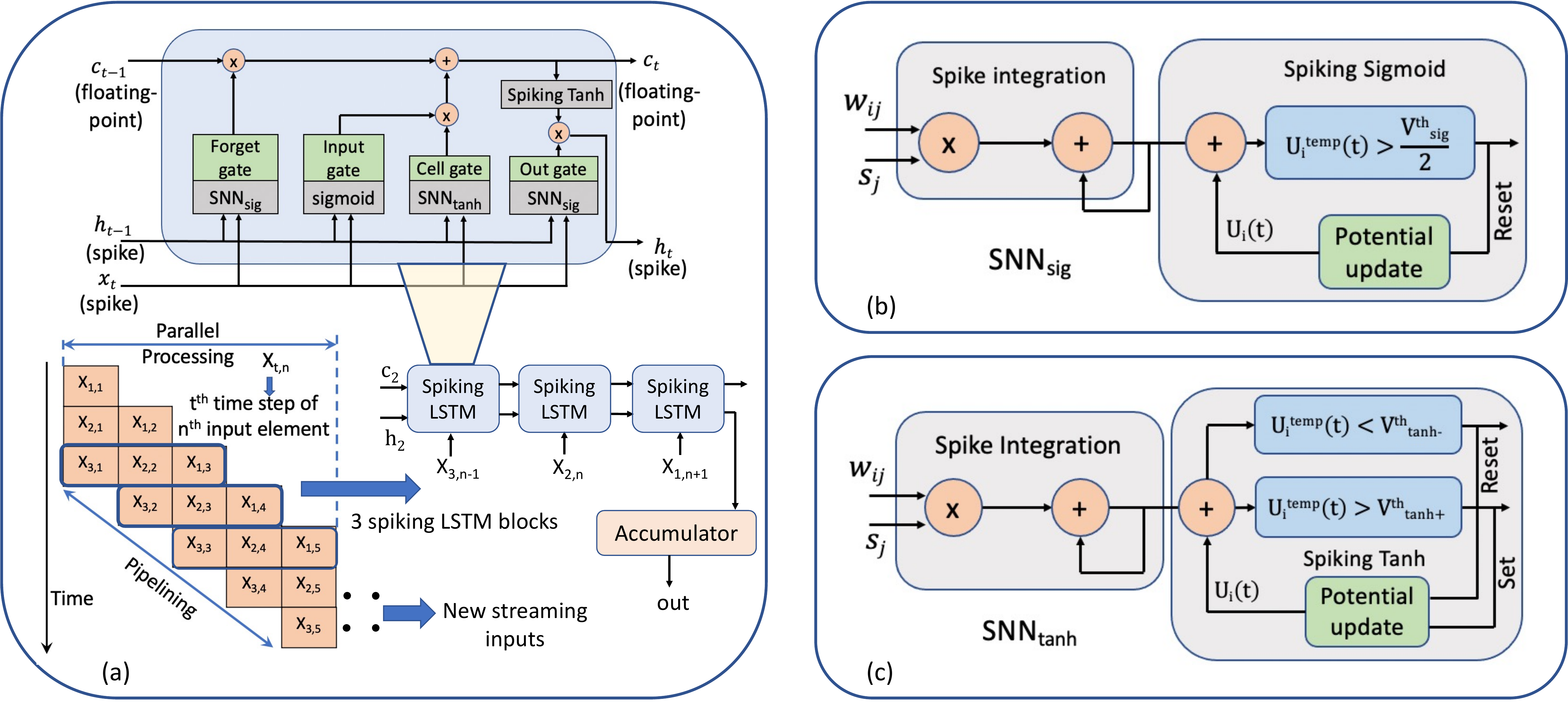}
\end{center}
\caption{(a) Proposed spiking LSTM architecture and dataflow with the parallel pipelined execution for the example of 5 time steps and 3 input elements in the sequence. LIF activation function corresponding to the (b) sigmoid and (c) tanh activation function used in the spiking LSTM architecture.}
\vspace{-4mm}
\label{fig:spiking_LSTM_dataflow}
\end{figure*}

The LIF outputs $S_{sig}(t)$ and $S_{tanh}(t)$ at time step $t$ converted from the sigmoid and tanh functions respectively is 
\vspace{-2mm}
\begin{align}\label{eq:snn_activation_sig}
   S_{sig}(t)&=
\begin{dcases}
    1, & \text{if } U_{sig}^{temp}(t)>\frac{V_{sig}^{th}}{2} \\
    0, & \text{otherwise}, \\
\end{dcases} \\
\qquad 
S_{tanh}(t) &=
\begin{dcases}
    1, & \text{if } U_{tanh}^{temp}(t)>{V_{tanh+}^{th}} \\
    -1, & \text{if } U_{tanh}^{temp}(t)<{V_{tanh-}^{th}} \\
    0, & \text{otherwise}, 
\end{dcases}
\end{align}
as illustrated in Fig. \ref{fig:spiking_LSTM_dataflow}(b-c). Note that 
$ U_{sig}^{temp}(t)$ and $ U_{tanh}^{temp}(t)$ denote the accumulated membrane potentials at time step $t$ and that we compare with $\frac{V_{sig}^{th}}{2}$ for the hard sigmoid activation as the curve is symmetric around $0$.

Given a pre-trained baseline LSTM model, our first objective is to minimize the conversion error between the baseline and spiking model. Inspired by the work in \cite{deng2021optimal} that focuses on CNNs, we achieve this objective by minimizing the difference between the outputs of the non-spiking LSTM activation functions and the IF activation functions (that model the average IF activation outputs over all time steps, as shown below, where $\Bar{{X}}_{sig}$ and $\Bar{{X}}_{tanh}$ denote the time-averaged inputs).\footnote{The proof that these functions captures the average is presented in Appendix A.}
\vspace{-1.5mm}
\begin{align}
    \Bar{Y}_{sig} &= \frac{1}{T}\text{clip}\left(\floor{\frac{T}{V^{th}_{sig}}\left({W}\Bar{{X}}_{sig}+\frac{V^{th}_{sig}}{2}\right)},0,T\right) \notag
\end{align}
\vspace{-1.5mm}
\begin{align}
\Bar{Y}_{tanh} &=
\begin{cases}
    \frac{1}{T}\text{clip}\left(\floor{\frac{T}{V^{th}_{tanh+}}{W}\Bar{{X}}_{tanh}},0,T\right), \text{if } \textit{A} \\
    \frac{1}{T}\text{clip}\left(\floor{\frac{T}{V^{th}_{tanh-}}{W}\Bar{{X}}_{tanh}},-T,0\right), \text{otherwise} \notag \\
\end{cases}
\end{align}
where \textit{A} corresponds to ${W}\Bar{{X}}_{tanh}>0$.
As illustrated in Fig. \ref{fig:act_func}(a), the IF activation output is always less than the sigmoid counterpart, and hence the error accumulates over the multiple time steps and input elements in the sequence. To mitigate this error, the IF activation curve needs to be shifted to the left, which can be done by adding a bias term as shown in Fig. \ref{fig:act_func}(a). Under the assumption that the inputs to the IF and hard sigmoid functions are uniformly and identically distributed (IID), the optimal value of this bias is $V^{th}_{sig}/2T$. Similarly, for the tanh function, we require a bias addition of $V^{th}_{tanh+}/2T$ when the input is positive and a bias subtraction of $V^{th}_{tanh-}/2T$ when the input is negative (both proofs are shown in Appendix B). Reference \cite{deng2021optimal} also proposed an optimal bias term for the ReLU activation used for static image recognition tasks, which the authors claim helps reduce the number of time steps. However, a follow-up work \cite{datta2022date} observed a significant accuracy drop for ultra low time steps ($2{-}4$) even with this optimal shift. They claimed this is because of the flawed IID assumption and that the ReLU activation inputs are highly skewed towards $0$. However, for the sigmoid and tanh activations, we observe a relatively even distribution around $0$, as shown in Appendix B and hence, for ultra low time steps, our proposed shifts still yield decent test accuracy, as shown in Table \ref{tab:main_results}.
\vspace{-1mm}

\subsection{SNN Training}

Once we initialize these IF activation functions, we aim to further optimize the error between the outputs of the IF and non-spiking LSTM activations in order to reduce the number of time steps and resulting energy consumption. Note that this error stems from the non-IID distributions of the tanh and sigmoid inputs which are hard to model. We convert the IF model to its LIF counterpart by incorporating the leak term that provides a tunable control knob that can be leveraged to minimize this error. To further motivate this argument, let us derive the LIF activation function converted from the hard sigmoid function. Assuming the initial membrane potential of any neuron $U_i(t)=0$, substituting $U_i^{temp}(t)$ in Eq. (1), and summing over $T$ time steps, $U_i(t)$ can be expressed as
\vspace{-1mm}
\begin{align}
(1{+}\lambda{+}\lambda^2{+}..\lambda^{t{-}1})\sum_j{W_{ij}\Bar{X}_{j}}{=}\left(\frac{1{-}\lambda^t}{1{-}\lambda}\right)\sum_j{W_{ij}\Bar{X}_{j}} \notag
\end{align}
\vspace{0mm}
Assuming the $i^{th}$ neuron first spikes at time $t$, $U_i(t)>V^{th}_{sig}$ and  $U_i(t-1)<V^{th}_{sig}$. These two conditions respectively imply
\vspace{-1.5mm}
\begin{align*}
& \left(\frac{1-\lambda^t}{1-\lambda}\right)\sum_j{W_{ij}\Bar{X}_{j}}>V^{th}_{sig}
 \Rightarrow{t}{>}\frac{log\left(1{-}\frac{V^{th}_{sig}(1{-}\lambda)}{\sum_j{W_{ij}\Bar{X}_{j}}}\right)}{log(\lambda)} \\
& \left(\frac{1{-}{\lambda}^{t{-}1}}{1{-}\lambda}\right)\sum_j{W_{ij}\Bar{X}_{j}}{<}V^{th}_{sig}\Rightarrow{t}{<}1{+}\frac{log\left(1{-}\frac{V^{th}_{sig}(1-\lambda)}{\sum_j{W_{ij}\Bar{X}_{j}}}\right)}{log(\lambda)}
\end{align*}

\noindent
Note that if ${\sum_j{W_{ij}\Bar{X}_{j}}}<V^{th}_{sig}(1-\lambda)$, the $i^{th}$ neuron will never spike, and hence, the average output activation will be $0$. The above two inequalities imply
\vspace{-1mm}
\begin{align}\label{eq:snn_activation_time}
t=\ceil{\frac{log\left(1-\frac{V^{th}_{sig}(1-\lambda)}{\sum_j{W_{ij}\Bar{X}_{j}}}\right)}{log(\lambda)}}.
\end{align}
Assuming the membrane potential resets to $0$ when it crosses the threshold, the $i^{th}$ neuron spikes $\floor{\frac{T}{t}}$ times. Hence, the LIF output can be estimated as 
\vspace{-2mm}
\begin{align}\label{eq:modified_lif_2}
 \Bar{Y}_{sig} =
\begin{cases}
     \frac{1}{T}\floor{\frac{T}{t}},  & \text{if }  {{W\Bar{X}_{sig}}}>V^{th}_{sig}(1-\lambda) \\
   0, & \text{otherwise}, \\
\end{cases}
\end{align}
which has been illustrated in Fig. \ref{fig:act_func}(a) for two different cases, one for which $\lambda>1$ and one for which $\lambda<1$. The optimal value of $\lambda$ can help reduce the difference between the LIF and non-spiking LSTM output activations, and it depends on the nature of the two input distributions \rev{which are difficult to model}. Note that $\Bar{Y}_{sig}$ in Eq. (7) also depends on $V^{th}_{sig}$ which provides another tunable control knob for error minimization. Hence, \rev{motivated by the derivation leading to Eq. (7)} we optimize the threshold and leak term via SGL during SNN training, along with the weights. \rev{Note that this derivation might also be useful for the SNN community to model discrete neuron activation functions, thereby helping to bridge the gap between deep learning and neuromorphic computing.}

\begin{table*}
\scriptsize\addtolength{\tabcolsep}{-1.0pt}
  \caption{Test accuracy on temporal MNIST, GSC, and UCI datasets obtained by proposed approaches with direct encoding for 2 time steps. S and NS denote the spiking and non-spiking LSTM models respectively. On the other hand, P and NP denotes the accuracies with and without a pre-trained non-spiking LSTM model respectively.}
  \label{tab:main_results}
  \centering
  \begin{tabular}{llllllllllll}
    \toprule
    \cmidrule(r){1-12}
    LSTM      & $V^{th}$  & $V^{th}$  & $\lambda$  & NS & NS  & \multicolumn{2}{c}{T-MNIST Acc. (\%)} & \multicolumn{2}{c}{GSC Acc. (\%)} & \multicolumn{2}{c}{UCI Acc. (\%)} \\
    \cmidrule(r){7-12}
    Model  & Shift & Train & Train & sig$_i$ & tanh$_g$ & \centered{P} & \centered{NP} & \centered{P} & \centered{NP} & \centered{P} & \centered{NP} \\
    \midrule
        \centered{NS}     & \centered{\xmark} & \centered{\xmark} & \centered{--} & \centered{--} & \centered{--} & \centered{--}  &\centered{98.6{$\pm$}0.2} &\centered{--} & \centered{95.42{$\pm$}0.1}& \centered{--} & \centered{90.37{$\pm$}0.2}  \\
        \cmidrule(r){1-12}
   \centered{S}     & \centered{\xmark}  & \centered{\xmark} & \centered{\xmark} & \centered{\xmark} & \centered{\xmark} &\centered{97.84{$\pm$}0.2}  &\centered{97.74{$\pm$}0.3} & \centered{90.59{$\pm$}0.2} & \centered{63.45{$\pm$}0.2} & \centered{88.17{$\pm$}0.2} &  \centered{87.63{$\pm$}0.1} \\ 
    & \centered{\cmark}  &  \centered{\xmark} & \centered{\xmark} & \centered{\xmark} & \centered{\xmark} & \centered{97.98{$\pm$}0.2} & \centered{97.87{$\pm$}0.1} & \centered{92.05{$\pm$}0.1}& \centered{91.45{$\pm$}0.2} & \centered{88.60{$\pm$}0.2} & \centered{88.13{$\pm$}0.3}  \\  
    & \centered{\cmark}  & \centered{\cmark} & \centered{\xmark} & \centered{\xmark} & \centered{\xmark} & \centered{97.92{$\pm$}0.1} & \centered{97.84{$\pm$}0.2} & \centered{92.87{$\pm$}0.2}& \centered{91.33{$\pm$}0.3} & \centered{88.64{$\pm$}0.3} & \centered{86.87{$\pm$}0.2}  \\ 
    & \centered{\cmark}  & \centered{\cmark} & \centered{\cmark} & \centered{\xmark} & \centered{\xmark} & \centered{98.0{$\pm$}0.2}  & \centered{97.95{$\pm$}0.2} & \centered{93.57{$\pm$}0.1}&\centered{92.14{$\pm$}0.1} & \centered{89.13{$\pm$}0.4} & \centered{87.50{$\pm$}0.3}  \\ 
    & \centered{\cmark}  & \centered{\cmark} & \centered{\cmark} & \centered{\cmark} & \centered{\xmark} &\centered{98.1{$\pm$}0.3}  & \centered{97.98{$\pm$}0.1} &\centered{94.75{$\pm$}0.1} & \centered{92.63{$\pm$}0.2} & \centered{89.23{$\pm$}0.2} & \centered{89.20{$\pm$}0.2}  \\ 
    & \centered{\cmark}  & \centered{\cmark} & \centered{\cmark} & \centered{\xmark} & \centered{\cmark} & \centered{98.15{$\pm$}0.1}  & \centered{98.12{$\pm$}0.2} &\centered{94.53{$\pm$}0.2} & \centered{92.61{$\pm$}0.3} & \centered{89.40{$\pm$}0.3} & \centered{89.12{$\pm$}0.1}  \\ 
    \bottomrule
  \end{tabular}
  \vspace{-4mm}
\end{table*}

\subsection{Selective conversion of LSTM activation functions}

Instead of converting all the sigmoid and tanh activation functions in the non-spiking LSTM architecture to spiking counterparts, we judiciously select a subset of them such that we only need spike-based AC operations. This avoids the unnecessary accumulated error due to the spiking gradients and improves the inference accuracy at low time steps. To motivate our decision, let us review the equations governing the LSTM architecture in Eq. (8-10) where $h_t$ and $c_t$ denote the hidden and cell state tensors respectively. We denote $f_t$, $i_t$, $g_t$, and $o_t$ as outputs of the forget, input, cell, and output gates respectively. All weight tensors $w_{a,h}$, $w_{a,x}$, $w_{g,h}$, $w_{g,x}$ are assumed to be multi-bit values which is standard in SNN setups \cite{rathi2020iclr}.    
\vspace{-0.5mm}
\begin{align}\label{eq:snn_activation_lstm}
   \textbf{a}_t&{=}\text{sig}_{\textbf{a}}(w_{\textbf{a},h}h_{t-1}{+}w_{\textbf{a},x}x_t) \ \forall \textbf{a}{\in} \{f,i,o\}, \   \\
   g_t&{=}\text{tanh}_g(w_{g,h}h_{t-1}{+}w_{g,x}x_t) \\
   c_t&=f_t\odot c_{t-1}+i_t\odot g_t, \ h_t=o_t\odot \text{tanh}_c(c_t)
\end{align}
We propose to encode $x_t$ using spike tensors, as otherwise the MAC operation with the weight tensors would require costly multiplications. Similarly, $h_t$ also needs to be a spike tensor, which implies that $o_t$ should be a spike tensor and tanh$_c$ should be converted to a LIF activation. A spiking $o_t$ necessitates conversion of the sig$_o$ to LIF (see Eq. (9)). On the other hand, Eq. (10) implies that either $f_t$ or $c_{t-1}$ and $i_t$ or $g_t$ need to be a spike tensor in order to maintain multiplier-less operation. Between $f_t$ and $c_{t-1}$, we choose $f_t$ as the spike tensor because sig$_f$ can be easily converted to LIF activation as proposed in this work and because $c_{t-1}$ is the sum of two tensors which is not naturally a spike tensor. Moreover, in the LSTM architecture there is no activation function applied to the floating point tensor $c_{t-1}$. Thus, $c_{t-1}$ cannot be converted to a spike tensor. Between $i_t$ and $g_t$, we can arbitrarily convert either to a spike tensor (for illustration purposes, we convert $i_t$ in Fig. \ref{fig:spiking_LSTM_dataflow}). 

\vspace{-2mm}
\section{Pipelined Parallel SNN Processing}

Let us denote the length of the input sequence to be processed by the spiking LSTM as $N$ and \rev{the total number of time steps over which each input spike tensor is encoded to approximate the original multi-bit inputs as $T$.}
To hide the latency incurred by the temporal dynamics of the LIF model, we propose a pipelined and parallel processing scheme depicted in Fig. \ref{fig:spiking_LSTM_dataflow}(a) for $N=5$ and $T=3$. \rev{In our scheme, the LSTM state is updated for every time step when each input element tensor is encoded to a spike tensor, modulated by the weights, and processed by the IF/LIF activation function. This allows us to start processing the next input tensor in the sequence, provided we have enough hardware resources or the resources for each LSTM block are internally pipelined. Note that we can process a maximum of $T$ input elements at the same time, which implies that for small $T$ (as shown in this work), the hardware overhead may be manageable. The state spike tensors $h_t$ and $c_t$ obtained from the first input element will continue to get updated once per time step until $T$ time steps. Since in each time step, a new input element in the sequence begins to be processed, the first spike input of the $N^{th}$ input element will be processed at the $N^{th}$ time step. To process the remaining $(T{-}1)$ spike inputs of its encoding, we need an additional $(T{-}1)$ time steps. Hence, the total number of time steps required to process the whole input sequence with our spiking LSTM is $(T{+}N{-}1)$.} For hardware with built-in parallel processing capability such as GPUs, our approach  improves the hardware utilization compared to non-spiking LSTMs that are sequential in nature. Note that previous research on LSTM-based SNNs \cite{ponghiran2021hyrbidsnn} accumulates the spike outputs of the different gates over all the time steps for processing a single input element. \rev{As a result, it uses $T{\times}N$ time steps to process the entire input sequence.} Moreover, the hidden state input to the next unrolled LSTM block becomes a multi-bit value which necessitates the use of energy-hungry multiplications.

\vspace{-2mm}

\vspace{-.5mm}
\section{Experimental Results}
We validate our proposed techniques on temporal MNIST \cite{lecun1998mnist}, Google speech commands (GSC) with 11 classes \cite{gsc}, and UCI smartphone datasets \cite{Anguita2013APD}. For temporal MNIST (T-MNIST), we use row-wise sequential inputs, resulting in $32$ image pixels each over a sequence of $32$ frames \cite{ponghiran2021hyrbidsnn, boijan2021spiking}. 
For GSC, we pre-process the raw audio inputs using log-mel spectrograms resulting in $20$ frequency features over a sequence of $81$ frames \cite{jeffares2022spikeinspired}. For UCI smartphone, we pre-process the sensor signals obtained from the smartphone worn on the waist of the participating humans by applying butterworth low-pass
filters within a fixed-width sliding windows of $2.56$ seconds
and $50\%$ overlap ($128$ readings per window) \cite{yu2018wcp}. For all the three datasets, we use both one and two-layer LSTMs with 128 hidden neurons in each layer. While we use a single fully-connected (FC) classifier layer for the T-MNIST and UCI datasets, we use two FC layers of 32 and 11 neurons each, with softmax output for the GSC dataset, \rev{following \cite{jeffares2022spikeinspired}}. We do not convert the FC layers to spiking counterparts as they consume ${<}{0.03\%}$ of total energy.  


\begin{figure*}[t!]
\begin{center}
\includegraphics[width=0.9\textwidth]{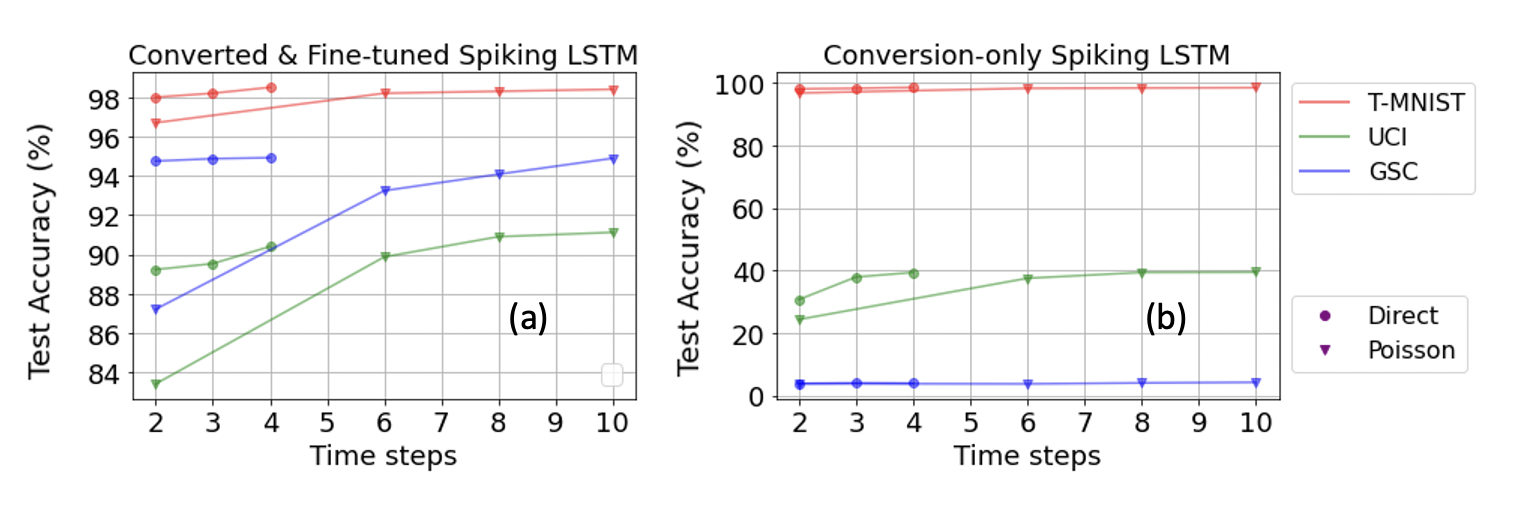}
\end{center}
\vspace{-4mm}
\caption{\rev{Comparison between the test accuracies obtained by our direct and Poisson encoded spiking LSTM models (a) with both conversion and SNN fine-tuning and (b) with only conversion.}}
\label{fig:direct_vs_Poisson_timesteps}
\vspace{-4.5mm}
\end{figure*}



\subsection{Inference Accuracy}

\textbf{Model Ablation}: Our results for single layer LSTMs are illustrated in Table \ref{tab:main_results} for $2$ time steps, both with and without a pre-trained non-spiking LSTM model. Each of our proposed techniques improve the test accuracy for large-scale tasks such as GSC, with an overall improvement of $3.7\%$. For T-MNIST, which is an easier task with less room to improve accuracy, we observe that our techniques lead to a $0.31\%$ improvement in accuracy, while the threshold and leak optimizations yield hardly any benefits. For the relatively more challenging UCI dataset, the leak optimization leads to the maximum accuracy increase ($+0.49\%$) with a pre-trained model, while the total increase due to all our techniques is $1.23\%$. While the UCI accuracy can be further increased with the use of bi-directional and stacked LSTMs \cite{har} ($+1.5\%$), it increases the energy. 
Note that the accuracies obtained without pre-trained models are lower; the difference with those with pre-trained models increases with the application complexity.

We also compare the impact of direct and Poisson encoding on the SNN accuracy in Fig. \ref{fig:direct_vs_Poisson_timesteps}(a-b). Note that the test accuracies obtained after only ANN-to-SNN conversion shown in Fig. \ref{fig:direct_vs_Poisson_timesteps}(b) are significantly lower than those obtained after SNN training, especially for more complex tasks. In particular, our approach, with SNN training yields close to state-of-the-art (SOTA) accuracy with only $2$ time steps, providing more than $7\times$ reduction in the latency compared to our conversion-only approach for the GSC dataset. Our conversion framework provides the optimal value of the SNN threshold (before SNN fine-tuning) backed by our theoretical insights. Thus, our conversion framework acts as a good initializer for the weights and the membrane potential, without which the test accuracy of more complex tasks, such as GSC, drops by ${>}3.4\%$. Note that our conversion framework alone also outperforms the existing ANN-to-SNN conversion frameworks for LSTMs. For example, our conversion-only spiking LSTM models yield $94.3\%$ test accuracy with $15$ time steps; the conversion-only baseline spiking models (without our proposed threshold shifts) require $32$ time steps to obtain similar accuracy. On the other hand, the baseline models can attain only 86\% accuracy with $15$ time steps.


\textbf{Comparison with prior works}: We compare the test accuracies obtained by our training framework with that of existing works in Table \ref{tab:snn_comparison}. Our accuracies are close to SOTA (within $0.3\%$) obtained by the non-spiking models, while yielding significant energy and latency savings as shown in \ref{fig:energy_and_delay}. We surpass the SOTA spiking models in terms of accuracy for both the T-MNIST and GSC datasets.\footnote{Note that we were unable to find any deep SNN architectures, classifying the UCI dataset for comparison.}

\renewcommand{\arraystretch}{1.0}
\begin{table*}
\caption{Accuracy comparison of the best performing models obtained by our training framework with state-of-the-art spiking and non-spiking LSTM models on T-MNIST and GSC dataset.}
\label{tab:snn_comparison}
\centering
\begin{tabular}{lllll}
\hline
Authors & Model & Training technique & Architecture & Accuracy (\%)   \\
\hline
 \multicolumn{5}{|c|}{Dataset : temporal MNIST} \\
\hline
\hline
\cite{neurips2017_cortical}  &  Spiking & BPTT & LSTM(128) & 97.29  \\
\hline
\cite{neurips_2018_bellec} & Spiking & BPTT & LSTM(220) & 96.4  \\
\hline
\cite{rezaabad2020icons} & Spiking & BPTT  & LSTM(1000)   & 98.23 \\
\hline
\cite{ponghiran2021hyrbidsnn} & Spiking & ANN-to-SNN conv. & LSTM(128)  & 98.72 (T=64) \\
\hline
\cite{boijan2021spiking} & Spiking & BPTT & RNN(64-256-256) & 98.7 \\
\hline
\cite{arjovsky2016rnn} & Non-spiking & BPTT & u-RNN(128) & 98.2 \\
\hline
\cite{jeffares2022spikeinspired} & Non-spiking & RC-BPTT & LSTM (320) & 98.14\\ 
\hline
This work & Spiking & Conv.+SNN training & LSTM(128) & \textbf{98.93 (T=8)} \\
\hline
\hline
 \multicolumn{5}{|c|}{Dataset : GSC} \\
\hline
\hline
\cite{morales2018scnn} & Spiking & BPTT & RNN(300-300) & 92.2 \\
\hline
\cite{pellegrini2021speech} & Spiking & BPTT & CNN(64-64-64) & 94.5 \\
\hline
\cite{jeffares2022spikeinspired} & Non-spiking & RC-BPTT & LSTM(128)  & \textbf{95.2} \\
\hline
\rev{\cite{salaj2020}} & \rev{Spiking} & \rev{BPTT}   & \rev{LSTM(128)} & \rev{91.2} \\
\hline
This work & Spiking & Conv.+SNN training &  LSTM(128) & \textbf{95.02 (T=4)} \\
\hline
\hline
 \multicolumn{5}{|c|}{\rev{Dataset : UCI}} \\
\hline
\hline
\cite{pellegrini2021speech} & \rev{Non-Spiking} & \rev{SGD} & \rev{Bi-dirLSTM(-)} & \rev{\textbf{91.1}} \\
\hline
\rev{This work} & \rev{Spiking} & \rev{Conv.+SNN training} &  \rev{LSTM(128)} & \rev{\textbf{90.78 (T=4)}} \\
\hline
\end{tabular}
\end{table*}

\vspace{-1mm}
\subsection{Inference Energy Efficiency}

\begin{figure*}[t!]
\begin{center}
\includegraphics[width=0.8\textwidth]{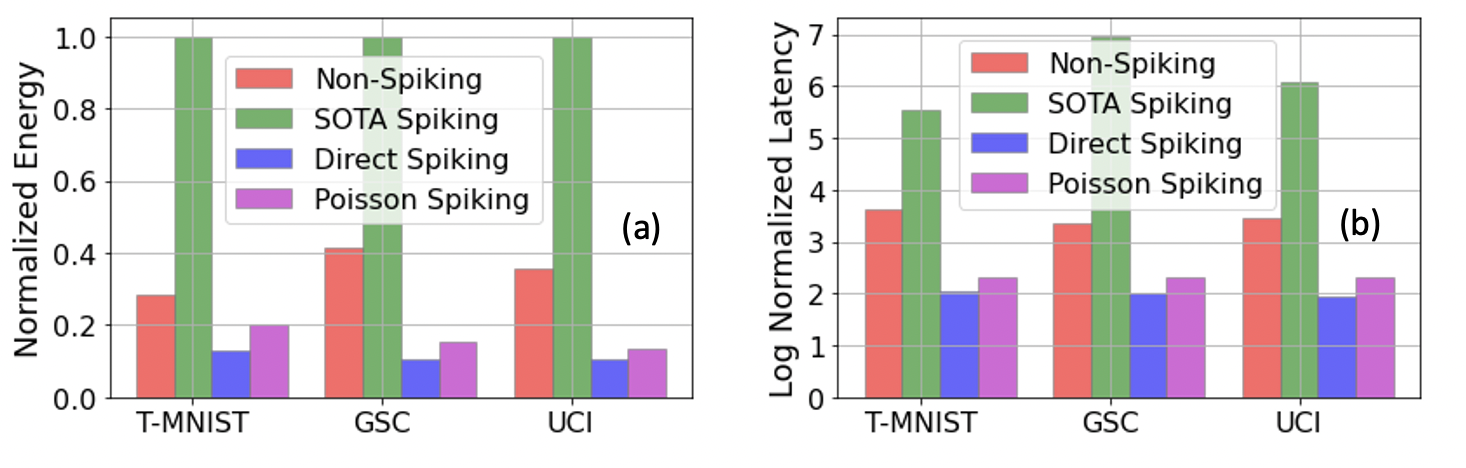}
\end{center}
\vspace{-4mm}
\caption{\rev{(a) Energy and (b) Delay comparisons between the non-spiking LSTM, proposed direct and Poisson encoded spiking LSTM, and the SOTA spiking LSTM model \cite{ponghiran2021hyrbidsnn}, that does not include any of our proposed approaches. }}
\label{fig:energy_and_delay}
\vspace{-2mm}
\end{figure*}

The inference compute energy is dominated by the total number of floating point operations (FLOPs). For non-spiking LSTMs, it consists of the MAC, AC, hard sigmoid and hard tanh operations required in the four gates. 
On the contrary, for spiking LSTMs, each emitted spike indicates which weights need to be accumulated at the post-synaptic neurons and results in a fixed number of AC operations. This, coupled with the comparison operations for the membrane potential in each time step dominates the SNN energy. 
\rev{We use custom RTL specifications and $28$ nm Kintex-7 FPGA platform to estimate the post place-and-route energy consumption of the hardware implementations of the spiking and non-spiking networks. In particular, we develop Verilog RTL block-level models to design, simulate, and synthesize an inference pipeline that captures the LSTM processing excluding the writing and reading of the weights and membrane potentials (which is dependent on the underlying micro-architecture and dataflow that is not modeled in our work) for the spiking LSTMs on our target FPGA device. In addition, for comparison purposes, we develop a similar synthesizable RTL design for the non-spiking LSTMs. Note, however, that our spiking LSTM designs can be further optimized by zero and clock gating that can help leverage sparsity and reduce computations, further reducing the energy.}
Fig. \ref{fig:energy_and_delay}(a) and (b) illustrate the energy consumption for our spiking and non-spiking LSTM architectures used for classifying the three datasets, along with the SOTA spiking LSTM implementation \cite{ponghiran2021hyrbidsnn}. As we can see, we obtain \rev{$2.8$-$5.1\times$} and \rev{$10.1$-$13.2\times$} lower energy than the non-spiking and SOTA spiking implementations respectively for direct coding. The \rev{reductions} obtained by Poisson encoding are a little lower ($1.8$-$3.5\times$ compared to non-spiking and $6.6$-$9.0\times$ compared to SOTA spiking) due to the degraded trade-off between more time steps and less energy due to ACs. 

On custom neuromorphic architectures, such as TrueNorth \cite{Merolla2014AMS}, and SpiNNaker \cite{spinnaker}, the total energy is estimated as $FLOPs{*}{E_{compute}}{+}{T}{*}{E_{static}}$ \cite{park_2020}, where the parameters $(E_{compute},E_{static})$ can be normalized to $(0.4, 0.6)$ and $(0.64, 0.36)$ for TrueNorth and SpiNNaker, respectively \cite{park_2020}. Since the total FLOPs for the LSTM architectures used in this and prior works \cite{jeffares2022spikeinspired,ponghiran2021hyrbidsnn} are several orders of magnitude higher than $T$, we would see similar compute energy improvements on them.

\vspace{-2mm}
\subsection{Inference Latency}
\rev{The latency incurred by the non-spiking LSTM architecture depends on
 the latency of our RTL-implemented MAC, AC, multiplication, hard sigmoid and hard tanh modules. Depending on the number of these units available in each LSTM block, we parallelize the MAC operations, followed by the AC and activation functions in the four LSTM gates. We incur additional AC and hard tanh delay to produce $c_t$ and $h_t$ respectively. In contrast, with our proposed implementation, both the direct and Poisson encoded spiking LSTM architectures can process $T$ different input elements simultaneously by internally pipelining the RTL models.}
Note that we use the similar RTL models and FPGA evaluation setup illustrated above to evaluate the latency of the LSTM implementations. As shown in Fig. \ref{fig:energy_and_delay}(b), our processing scheme, coupled with the increased sparsity and accumulate-only operations in SNN, results in \rev{${\sim}4\times$ and $25.9$-$105.7\times$} reduction in latency compared to the non-spiking LSTM and SOTA spiking LSTM implementation respectively. \rev{This significant improvement over the SOTA spiking implementation can be attributed to three factors. Firstly, the SOTA spiking LSTM architecture require more time steps ($3${-}$8\times$) to encode the original multi-bit input tensor than ours to achieve similar test accuracy. Secondly, while our proposed spiking architecture requires a total of ($T{+}N{-}1$) time steps to process the whole input sequence, the existing spiking counterpart requires $T'\times N$ time steps where $T$ and $T'$ denote the total number of SNN time steps for the proposed and existing networks respectively. Lastly, since the hidden and cell state tensors are multi-bit tensors, the LSTM block requires MACs for certain computations, which also increases the latency by $5.1\times$ obtained from FPGA simulations compared to our AC-only approach.}
\vspace{-2mm}
\section{Conclusions \& Broader Impact}
\vspace{-1mm}
In this work, we propose a \rev{spiking LSTM training framework} which significantly reduces the inference latency and energy efficiency compared to existing works with minimal ($<0.3\%$) accuracy drop for diverse large-scale streaming ML use cases. ML models for large-scale streaming tasks are typically compute-intensive and therefore need cloud hardware with a lot of processing power that can massively increase the environmental carbon footprint \cite{MITenvironmental}.
Our models can reduce this footprint by allowing smart home assistants and wearable sensors with small compute and memory footprints to perform on-device audio and action recognition respectively. We can also enable real-time inference, thereby improving the user experience. 
While our goal is to enable socially responsible applications, our work can also enable
cheap and real-time speech recognition systems that might be susceptible to adversarial attacks. Similarly, our technology can also be abused in certain sensor wearables. 
Preventing the application of this technology from abusive usages is an important and interesting area of future work. 

\section{Acknowledgements}

We would like to acknowledge the DARPA HR$00112190120$ award and the NSF CCF-$1763747$ award for supporting this work. The views and conclusions contained herein are those of the authors and should not be interpreted as necessarily representing the official policies or endorsements of DARPA or NSF.

{
\small

\bibliography{egbib}

\begin{thebibliography}{45}
\providecommand{\natexlab}[1]{#1}

\bibitem[{MIT(2019)}]{MITenvironmental}
 2019.
\newblock {Training a single AI model can emit as much carbon as five cars in
  their lifetimes}.
\newblock
  \url{https://www.technologyreview.com/2019/06/06/239031/training-a-single-ai-model-can-emit-as-much-carbon-as-five-cars-in-their-lifetimes/}.
\newblock Accessed: 06-06-2019.

\bibitem[{Anguita et~al.(2013)Anguita, Ghio, Oneto, Parra, and
  Reyes-Ortiz}]{Anguita2013APD}
Anguita, D.; Ghio, A.; Oneto, L.; Parra, X.; and Reyes-Ortiz, J.~L. 2013.
\newblock A Public Domain Dataset for Human Activity Recognition using
  Smartphones.
\newblock In \emph{21$^{st}$ European Symposium on Artificial Neural Networks,
  Computational Intelligence and Machine Learning}.

\bibitem[{Arjovsky, Shah, and Bengio(2016)}]{arjovsky2016rnn}
Arjovsky, M.; Shah, A.; and Bengio, Y. 2016.
\newblock Unitary Evolution Recurrent Neural Networks.
\newblock In \emph{Proceedings of the 33rd International Conference on
  International Conference on Machine Learning - Volume 48}, ICML'16,
  1120–1128. JMLR.org.

\bibitem[{Bellec et~al.(2018{\natexlab{a}})Bellec, Salaj, Subramoney,
  Legenstein, and Maass}]{bellec_2018long}
Bellec, G.; Salaj, D.; Subramoney, A.; Legenstein, R.; and Maass, W.
  2018{\natexlab{a}}.
\newblock Long short-term memory and learning-to-learn in networks of spiking
  neurons.
\newblock \emph{arXiv preprint arXiv:1803.09574}.

\bibitem[{Bellec et~al.(2018{\natexlab{b}})Bellec, Salaj, Subramoney,
  Legenstein, and Maass}]{neurips_2018_bellec}
Bellec, G.; Salaj, D.; Subramoney, A.; Legenstein, R.~A.; and Maass, W.
  2018{\natexlab{b}}.
\newblock Long short-term memory and Learning-to-learn in networks of spiking
  neurons.
\newblock In \emph{NeurIPS}, 795--805.

\bibitem[{{Comsa} et~al.(2020)}]{comsa_2020}
{Comsa}, I.~M.; et~al. 2020.
\newblock Temporal Coding in Spiking Neural Networks with Alpha Synaptic
  Function.
\newblock In \emph{ICASSP 2020 - 2020 IEEE International Conference on
  Acoustics, Speech and Signal Processing (ICASSP)}, volume~1, 8529--8533.

\bibitem[{Costa et~al.(2017)Costa, Assael, Shillingford, de~Freitas, and
  Vogels}]{neurips2017_cortical}
Costa, R.; Assael, I.~A.; Shillingford, B.; de~Freitas, N.; and Vogels, T.
  2017.
\newblock Cortical microcircuits as gated-recurrent neural networks.
\newblock In Guyon, I.; Luxburg, U.~V.; Bengio, S.; Wallach, H.; Fergus, R.;
  Vishwanathan, S.; and Garnett, R., eds., \emph{Advances in Neural Information
  Processing Systems}, volume~30.

\bibitem[{Datta and Beerel(2021)}]{datta2022date}
Datta, G.; and Beerel, P.~A. 2021.
\newblock Can Deep Neural Networks be Converted to Ultra Low-Latency Spiking
  Neural Networks?
\newblock \emph{arXiv preprint arXiv:2112.12133}.

\bibitem[{Datta et~al.(2022)Datta, Kundu, Jaiswal, and Beerel}]{datta2022fin}
Datta, G.; Kundu, S.; Jaiswal, A.~R.; and Beerel, P.~A. 2022.
\newblock {ACE-SNN:} Algorithm-Hardware Co-design of Energy-Efficient \&
  Low-Latency Deep Spiking Neural Networks for 3D Image Recognition.
\newblock \emph{Frontiers in Neuroscience}, 16.

\bibitem[{Datta et~al.(2021)}]{datta2021training}
Datta, G.; et~al. 2021.
\newblock Training Energy-Efficient Deep Spiking Neural Networks with
  Single-Spike Hybrid Input Encoding.
\newblock In \emph{2021 International Joint Conference on Neural Networks
  (IJCNN)}, volume~1, 1--8.

\bibitem[{Deng and Gu(2021)}]{deng2021optimal}
Deng, S.; and Gu, S. 2021.
\newblock Optimal Conversion of Conventional Artificial Neural Networks to
  Spiking Neural Networks.
\newblock In \emph{International Conference on Learning Representations}.

\bibitem[{Diehl et~al.(2016{\natexlab{a}})Diehl, Zarrella, Cassidy, Pedroni,
  and Neftci}]{diehl2016rnn_snn}
Diehl, P.~U.; Zarrella, G.; Cassidy, A.; Pedroni, B.~U.; and Neftci, E.
  2016{\natexlab{a}}.
\newblock Conversion of artificial recurrent neural networks to spiking neural
  networks for low-power neuromorphic hardware.
\newblock In \emph{2016 IEEE International Conference on Rebooting Computing
  (ICRC)}, volume~1, 1--8.

\bibitem[{Diehl et~al.(2016{\natexlab{b}})}]{diehl2016conversion}
Diehl, P.~U.; et~al. 2016{\natexlab{b}}.
\newblock Conversion of artificial recurrent neural networks to spiking neural
  networks for low-power neuromorphic hardware.
\newblock In \emph{2016 IEEE International Conference on Rebooting Computing
  (ICRC)}, 1--8. IEEE.

\bibitem[{Dominguez-Morales et~al.(2018)Dominguez-Morales, Liu, James,
  Gutierrez-Galan, Jimenez-Fernandez, Davidson, and Furber}]{morales2018scnn}
Dominguez-Morales, J.~P.; Liu, Q.; James, R.; Gutierrez-Galan, D.;
  Jimenez-Fernandez, A.; Davidson, S.; and Furber, S. 2018.
\newblock Deep Spiking Neural Network model for time-variant signals
  classification: a real-time speech recognition approach.
\newblock In \emph{2018 International Joint Conference on Neural Networks
  (IJCNN)}, volume~1, 1--8.

\bibitem[{Furber et~al.(2014)}]{spinnaker}
Furber, S.~B.; et~al. 2014.
\newblock The SpiNNaker Project.
\newblock \emph{Proceedings of the IEEE}, 102(5): 652--665.

\bibitem[{Jeffares et~al.(2022)Jeffares, Guo, Stenetorp, and
  Moraitis}]{jeffares2022spikeinspired}
Jeffares, A.; Guo, Q.; Stenetorp, P.; and Moraitis, T. 2022.
\newblock Spike-inspired rank coding for fast and accurate recurrent neural
  networks.
\newblock In \emph{International Conference on Learning Representations}.

\bibitem[{Kheradpisheh et~al.(2020)}]{Kheradpisheh_2020}
Kheradpisheh, S.~R.; et~al. 2020.
\newblock Temporal Backpropagation for Spiking Neural Networks with One Spike
  per Neuron.
\newblock \emph{International Journal of Neural Systems}, 30(06).

\bibitem[{Kim and Panda(2021{\natexlab{a}})}]{dynvis_1}
Kim, Y.; and Panda, P. 2021{\natexlab{a}}.
\newblock Optimizing Deeper Spiking Neural Networks for Dynamic Vision Sensing.
\newblock \emph{Neural Netw.}, 144(C): 686–698.

\bibitem[{Kim and Panda(2021{\natexlab{b}})}]{kim_2020}
Kim, Y.; and Panda, P. 2021{\natexlab{b}}.
\newblock Revisiting Batch Normalization for Training Low-Latency Deep Spiking
  Neural Networks From Scratch.
\newblock \emph{Frontiers in Neuroscience}, 15.

\bibitem[{{Kundu} et~al.(2021)}]{kundu_2021}
{Kundu}, S.; et~al. 2021.
\newblock Spike-Thrift: Towards Energy-Efficient Deep Spiking Neural Networks
  by Limiting Spiking Activity via Attention-Guided Compression.
\newblock In \emph{Proceedings of the IEEE/CVF Winter Conference on
  Applications of Computer Vision (WACV)}, 3953--3962.

\bibitem[{Kundu et~al.(2021)}]{kundu2021lowlatency}
Kundu, S.; et~al. 2021.
\newblock Towards Low-Latency Energy-Efficient Deep {SNNs} via Attention-Guided
  Compression.
\newblock \emph{arXiv preprint arXiv:2107.12445}.

\bibitem[{Lecun et~al.(1998)Lecun, Bottou, Bengio, and
  Haffner}]{lecun1998mnist}
Lecun, Y.; Bottou, L.; Bengio, Y.; and Haffner, P. 1998.
\newblock Gradient-based learning applied to document recognition.
\newblock \emph{Proceedings of the IEEE}, 86(11): 2278--2324.

\bibitem[{Lee et~al.(2020{\natexlab{a}})Lee, Kosta, Zhu, Chaney, Daniilidis,
  and Roy}]{lee_2020}
Lee, C.; Kosta, A.~K.; Zhu, A.~Z.; Chaney, K.; Daniilidis, K.; and Roy, K.
  2020{\natexlab{a}}.
\newblock Spike-FlowNet: Event-based Optical Flow Estimation with
  Energy-Efficient Hybrid Neural Networks.
\newblock arXiv:2003.06696.

\bibitem[{Lee et~al.(2020{\natexlab{b}})}]{leefin2020}
Lee, C.; et~al. 2020{\natexlab{b}}.
\newblock Enabling Spike-Based Backpropagation for Training Deep Neural Network
  Architectures.
\newblock \emph{Frontiers in Neuroscience}, 14.

\bibitem[{Lee et~al.(2016)}]{lee_dsnn}
Lee, J.~H.; et~al. 2016.
\newblock Training Deep Spiking Neural Networks Using Backpropagation.
\newblock \emph{Frontiers in Neuroscience}, 10.

\bibitem[{Li et~al.(2022)Li, Kim, Park, Geller, and Panda}]{dynvis_2}
Li, Y.; Kim, Y.; Park, H.; Geller, T.; and Panda, P. 2022.
\newblock Neuromorphic Data Augmentation for Training Spiking Neural Networks.
\newblock \emph{arXiv preprint arXiv:2203.06145}.

\bibitem[{Lotfi~Rezaabad and Vishwanath(2020)}]{rezaabad2020icons}
Lotfi~Rezaabad, A.; and Vishwanath, S. 2020.
\newblock Long Short-Term Memory Spiking Networks and Their Applications.
\newblock In \emph{International Conference on Neuromorphic Systems 2020},
  ICONS 2020. New York, NY, USA: Association for Computing Machinery.
\newblock ISBN 9781450388511.

\bibitem[{Merolla et~al.(2014)}]{Merolla2014AMS}
Merolla, P.; et~al. 2014.
\newblock A million spiking-neuron integrated circuit with a scalable
  communication network and interface.
\newblock \emph{Science}, 345: 668--673.

\bibitem[{Moritz, Hori, and Roux(2019)}]{Moritz2019UnidirectionalNN}
Moritz, N.; Hori, T.; and Roux, J.~L. 2019.
\newblock Unidirectional Neural Network Architectures for End-to-End Automatic
  Speech Recognition.
\newblock In \emph{INTERSPEECH}.

\bibitem[{{Neftci}, {Mostafa}, and {Zenke}(2019)}]{neftci_surg}
{Neftci}, E.~O.; {Mostafa}, H.; and {Zenke}, F. 2019.
\newblock Surrogate Gradient Learning in Spiking Neural Networks: Bringing the
  Power of Gradient-Based Optimization to Spiking Neural Networks.
\newblock \emph{IEEE Signal Processing Magazine}, 36(6): 51--63.

\bibitem[{Panda and Roy(2016)}]{panda2016_sup}
Panda, P.; and Roy, K. 2016.
\newblock Unsupervised Regenerative Learning of Hierarchical Features in
  Spiking Deep Networks for Object Recognition.
\newblock \emph{arXiv preprint arXiv:1602.01510}.

\bibitem[{Panda et~al.(2020)}]{panda_res}
Panda, P.; et~al. 2020.
\newblock Toward Scalable, Efficient, and Accurate Deep Spiking Neural Networks
  With Backward Residual Connections, Stochastic Softmax, and Hybridization.
\newblock \emph{Frontiers in Neuroscience}, 14.

\bibitem[{Park et~al.(2020)}]{park_2020}
Park, S.; et~al. 2020.
\newblock {T2FSNN}: Deep Spiking Neural Networks with Time-to-first-spike
  Coding.
\newblock \emph{arXiv preprint arXiv:2003.11741}.

\bibitem[{Pellegrini, Zimmer, and Masquelier(2020)}]{pellegrini2021speech}
Pellegrini, T.; Zimmer, R.; and Masquelier, T. 2020.
\newblock Low-activity supervised convolutional spiking neural networks applied
  to speech commands recognition.
\newblock \emph{arXiv preprint arXiv:2011.06846}.

\bibitem[{Ponghiran and Roy(2021{\natexlab{a}})}]{ponghiran2021hyrbidsnn}
Ponghiran, W.; and Roy, K. 2021{\natexlab{a}}.
\newblock Hybrid Analog-Spiking Long Short-Term Memory for Energy Efficient
  Computing on Edge Devices.
\newblock In \emph{2021 Design, Automation Test in Europe Conference Exhibition
  (DATE)}, volume~1, 581--586.

\bibitem[{Ponghiran and Roy(2021{\natexlab{b}})}]{ponghiran2021seqsnn}
Ponghiran, W.; and Roy, K. 2021{\natexlab{b}}.
\newblock Spiking Neural Networks with Improved Inherent Recurrence Dynamics
  for Sequential Learning.
\newblock \emph{arXiv preprint arXiv:2109.01905}.

\bibitem[{Rathi et~al.(2020{\natexlab{a}})Rathi, Srinivasan, Panda, and
  Roy}]{rathi2020iclr}
Rathi, N.; Srinivasan, G.; Panda, P.; and Roy, K. 2020{\natexlab{a}}.
\newblock Enabling Deep Spiking Neural Networks with Hybrid Conversion and
  Spike Timing Dependent Backpropagation.
\newblock In \emph{International Conference on Learning Representations}.

\bibitem[{Rathi et~al.(2020{\natexlab{b}})}]{rathi2020dietsnn}
Rathi, N.; et~al. 2020{\natexlab{b}}.
\newblock {DIET-SNN}: Direct Input Encoding With Leakage and Threshold
  Optimization in Deep Spiking Neural Networks.
\newblock \emph{arXiv preprint arXiv:2008.03658}.

\bibitem[{Rezaabad and Vishwanath(2020)}]{rezaabad2020lstmspike}
Rezaabad, A.~L.; and Vishwanath, S. 2020.
\newblock Long Short-Term Memory Spiking Networks and Their Applications.
\newblock \emph{arXiv preprint arXiv:2007.04779}.

\bibitem[{Salaj et~al.(2021)Salaj, Subramoney, Kraisnikovic, Bellec,
  Legenstein, and Maass}]{salaj2020}
Salaj, D.; Subramoney, A.; Kraisnikovic, C.; Bellec, G.; Legenstein, R.; and
  Maass, W. 2021.
\newblock Spike frequency adaptation supports network computations on
  temporally dispersed information.
\newblock \emph{eLife}, 10: e65459.

\bibitem[{Sengupta et~al.(2019)}]{dsnn_conversion_abhronilfin}
Sengupta, A.; et~al. 2019.
\newblock Going Deeper in Spiking Neural Networks: {VGG} and Residual
  Architectures.
\newblock \emph{Frontiers in Neuroscience}, 13: 95.

\bibitem[{Warden(2018)}]{gsc}
Warden, P. 2018.
\newblock Speech Commands: A Dataset for Limited-Vocabulary Speech Recognition.
\newblock \emph{arXiv preprint arXiv:1804.03209}.

\bibitem[{Yin, Corradi, and Bohte(2021)}]{boijan2021spiking}
Yin, B.; Corradi, F.; and Bohte, S.~M. 2021.
\newblock Accurate and efficient time-domain classification with adaptive
  spiking recurrent neural networks.
\newblock \emph{arXiv preprint arXiv:2103.12593}.

\bibitem[{Yu et~al.(2018)Yu, Chen, Yan, and Liu}]{yu2018wcp}
Yu, T.; Chen, J.; Yan, N.; and Liu, X. 2018.
\newblock A Multi-Layer Parallel LSTM Network for Human Activity Recognition
  with Smartphone Sensors.
\newblock In \emph{2018 10th International Conference on Wireless
  Communications and Signal Processing (WCSP)}, volume~1, 1--6.

\bibitem[{Zhao et~al.(2017)Zhao, Yang, Chevalier, and Gong}]{har}
Zhao, Y.; Yang, R.; Chevalier, G.; and Gong, M. 2017.
\newblock Deep Residual Bidir-LSTM for Human Activity Recognition Using
  Wearable Sensors.
\newblock \emph{arXiv preprint arXiv:1708.08989}.

\end{thebibliography}

}
\newpage



\end{document}